%% file: main.tex
\documentclass{article}

\PassOptionsToPackage{numbers,sort&compress}{natbib}
\usepackage[preprint]{neurips_2023}

\usepackage[utf8]{inputenc} 
\usepackage[T1]{fontenc}    
\usepackage{hyperref}       
\usepackage{url}            
\usepackage{booktabs}       
\usepackage{amsfonts}       
\usepackage{nicefrac}       
\usepackage{microtype}      
\usepackage{xcolor}         
\usepackage[normalem]{ulem} 
\usepackage{amsmath} 
\usepackage{cleveref}       
\usepackage{graphicx}  

\usepackage{algorithm}      
\usepackage{algpseudocode}  

\newcommand{\diff}{\mathrm{d}}
\renewcommand{\vec}[1]{\boldsymbol{#1}}

\newcommand{\Text}[2]{\item[]
\hspace*{-2em}
\noindent \textbf{#1}\ #2}

\title{Parallelized Acquisition for Active Learning using Monte Carlo Sampling}

\author{
  Jes\'{u}s Torrado\\
  Dipartimento di Fisica e Astronomia ``G. Galilei''\\
  Universit\`a degli Studi di Padova\\
  Via Marzolo 8, I-35131 Padova, Italy\\
  \texttt{jesus.torrado@pd.infn.it} \\
  \And
  Nils Sch\"oneberg\\
  Institut de Ci\`encies del Cosmos\\
  Universitat de Barcelona\\
  Mart\'{\i} i Franqu\`es 1, Barcelona E08028, Spain\\
  \texttt{nils.science@gmail.com} \\
  \And
  Jonas {El~Gammal}\\
  Department of Mathematics and Physics\\
  University of Stavanger\\
  NO-4036 Stavanger, Norway\\
  \texttt{jonas.e.elgammal@uis.no}
}

\usepackage[normalem]{ulem} 

\begin{document}

\maketitle

\begin{abstract}
Bayesian inference remains one of the most important tool-kits for any scientist, but increasingly expensive likelihood functions are required for ever-more complex experiments, raising the cost of generating a Monte Carlo sample of the posterior. Recent attention has been directed towards the use of emulators of the posterior based on Gaussian Process (GP) regression combined with active sampling to achieve comparable precision with far fewer costly likelihood evaluations. Key to this approach is the batched acquisition of proposals, so that the true posterior can be evaluated in parallel. This is usually achieved via sequential maximisation of the highly multimodal acquisition function. Unfortunately, this approach parallelizes poorly and is prone to getting stuck in local maxima. Our approach addresses this issue by generating nearly-optimal batches of candidates using an almost-embarrassingly parallel Nested Sampler on the mean prediction of the GP. The resulting nearly-sorted Monte Carlo sample is used to generate a batch of candidates ranked according to their sequentially conditioned acquisition function values at little cost. The final sample can also be used for inferring marginal quantities. Our proposed implementation (NORA) demonstrates comparable accuracy to sequential conditioned acquisition optimization and efficient parallelization in various synthetic and cosmological inference problems.
\end{abstract}

\section{Introduction}\label{sec:intro}

One of the fundamental tools of science is the comparison of observations with theory. In many Bayesian inference pipelines, this involves inferring the parameters of a model (or models themselves) given some observed or generated data. This is often realised directly using Bayes theorem: Given some model parameters $\vec{x}\in\mathbb{R}^d$ and data $\mathcal{D}$, the conditioned probability $p(\vec{x}|\mathcal{D})$ (the so-called \textit{posterior}) is given by
\begin{align}
    p(\vec{x}|\mathcal{D}) = \frac{p(\mathcal{D}|\vec{x})p(\vec{x})}{p(\mathcal{D})}\ .
\end{align}
where $p(\mathcal{D}|\vec{x})\equiv L(\vec{x})$ is called the \textit{likelihood}, $p(\vec{x})\equiv\pi(\vec{x})$ the \textit{prior}, and $p(\vec{D})\equiv E$ the \textit{evidence}. We are dropping the explicit dependence on $\mathcal{D}$ as it is fixed for a given inference problem. Traditionally the posterior distribution is sampled with Monte Carlo (MC) samplers such as Markov Chain Monte Carlo (MCMC) or Nested Sampling (NS).
Unfortunately though, $L(\vec{x})$ is often an expensive to evaluate black box function, either because calculating observables from the theoretical model involves expensive computations, because the amount of data is large, or both. This makes sampling in such circumstances unfeasible with MC samplers, since they typically require $\mathcal{O}(10^3-10^6)$ posterior evaluations for dimensionalities up to $\mathcal{O}(10)$.

There exist multiple approaches to accelerating such inference problems using machine-learning:
enhanced pre-conditioning to accelerate traditional MC methods \cite{Graff_2012,Moss:2019fyi,Hortua:2020exv,Williams:2021qyt,Karamanis:2022ksp},
simulation-based, implicit-likelihood inference algorithms \cite{Leclercq:2018who,Alsing:2019xrx,Miller:2020hua,Makinen:2021nly, Dai:2022dso, 2022mla..confE..52Z, Lemos:2022kua},
and emulators of underlying physical quantities (for Cosmological applications, see \cite{Manrique-Yus:2019hqc,Mootoovaloo:2020ott,SpurioMancini:2021ppk,To:2022ubu,Nygaard:2022wri,Gunther:2022pto}). In this work we are going to focus on emulating the likelihood as a function of its parameters, using a Gaussian Process (for previous approaches see
\cite{NIPS2012_6364d3f0,NIPS2014_e94f63f5,kandasamy:2015,2018arXiv180204782C,10.1162/neco_a_01127,Pellejero-Ibanez:2019enw,Gammal:2022eob} using GP, and \cite{NEURIPS2018_747c1bcc,NEURIPS2020_5d409541,huggins2023pyvbmc}
enhancing the GP with a variational approximation).
Any such emulation (such as that based on a GP) will typically require a set of samples of the function at various parameter points. In order to maximize the amount of information about the behavior of the function captured by these samples, often times an active sampling approach is used: New samples are proposed, based on the current best emulation, where the greatest probability of the estimated improvement of the future emulation is located \cite{NIPS2002_24917db1}. This is typically measured by an acquisition function. In this work we will tackle the question of how the active sampling algorithm can be performed in a highly parallel fashion while producing optimal or near-optimal batches of new proposed sampling locations. 


In order to acquire a nearly optimal batch of proposed sampling locations for the active learning algorithm in a highly parallel fashion, naive maximization of the acquisition function is not sufficient. This is not only due to the multi-modal nature of a typical acquisition function -- making it easy for the optimizer to get stuck in local maxima, especially in moderately high dimensionality -- but also due to the inherent lack of parallelization of standard optimization routines. This is caused by the sequential nature of the maximization algorithm, and more importantly requiring the result of a given maximization in order to compute the conditional acquisition function, which will be used for a subsequent maximization.

Our implementation combines the solution to both of these problems in an efficient way: First, by making use of a MC sampling algorithm it is possible to acquire samples of growing function value in a parallel fashion that is much more likely to find the global maximum. Second, through the usage of a ranked pool (see \Cref{sec:method}) we are also able to create a batch of multiple proposed sampling locations simultaneously. Both of these solutions combine to give us a highly efficient algorithm to acquire multiple near-optimal active learning sampling positions.

In \Cref{sec:method} we describe the general methodology employed in our algorithm. In \Cref{sec:results} we show the scaling with MPI processes as well as the acquisition histories for a number of toy examples, and we conclude in \Cref{sec:conclusion}. We also show further examples in the context of cosmological inference in \cref{app:sec:cosmo_examples}.

\section{Theoretical background}

\subsection{Gaussian Processes}
In this section we briefly summarize the main notation and theory. For a review, see \cite{gpml}. A Gaussian process (GP) is based on a probabilistic model of a function value at any point $x$, which follows a conditioned Gaussian with a mean $\mu$ and a standard deviation $\sigma$. Any two sampling locations $x$ and $x'$ are correlated in a multivariate Gaussian way with a correlation function given by $k(x,x')$, the kernel. The choice of the kernel and its hyperparameters encodes assumptions about the behavior of the underlying function (such as differentiability) into the GP. Given a set of sampling locations $X_1$,...,$X_N$ and corresponding function values $y_1$,...,$y_N$ the conditioned mean of the GP gives an emulation of the underlying function, while its conditional standard deviation describes the uncertainty in the emulation.
A common choice for the kernel function and the one that we will use throughout this paper is the radial basis function (RBF) kernel given in one dimension by
\begin{align}\label{eq:kernel_function}
    k(x,x') = C\cdot \exp\left(\frac{(x-x')^2}{2 l^2}\right)
\end{align}
where we will call $C$ the \textit{output scale} and $l$ the length scale of the kernel. In multiple dimensions ($\vec{x}\in\mathbb{R}^d$) we construct the kernel function as a product of RBF kernels, each acting on one dimension and have different length scales $l_i$:
\begin{align}\label{eq:kernel_multi}
    k(\vec{x},\vec{x}') = C\cdot \exp\left(\sum_{i=1}^d\frac{(x_i-x_i')^2}{2 l_i^2}\right)
\end{align}
By optimizing the marginal log-likelihood of the hyperparameters $\theta=\{C, \vec{l}\}$ one can fit the GP to a set of sampled points. A good choice of such samples, such as through an acquisition procedure for obtaining new locations is fundamental to the final performance of the GP emulation.

\subsection{Acquisition procedure}\label{sec:acq}
The second part, the acquisition of samples with which to train the GP relies on maximizing a so called \textit{acquisition function} which, given the current GP, is a measure of the assumed information gained by sampling at any given location. We will denote the already sampled point as the training set in this context. As we want more precision towards the top of the mode for the final inference steps 
we encode this by choosing the acquisition function
\begin{align}
   a(\mu, \sigma|\vec{x}) = 2\zeta(\mu(\vec{x})-p_{\mathrm{max}}) + \log(\sigma(\vec{x})) \ .
\end{align}
where $\zeta$ is an empirically determined dimensional regularization factor,\footnote{We use $\zeta = d^{-0.85}$, which has been shown in \cite{Gammal:2022eob} to provide a good balance between exploration and exploitation in a variety of dimensionalities.}  and $p_{\mathrm{max}}$ is the current maximum log-posterior value of the training set. We introduce this current maximum, as in most realistic cases the posterior distribution is not necessarily normalized and hence the scale of the peak not known.

In order to make use of the massive parallelization allowed for by current scientific computing systems, we require not a single optimal point, but a set of simultaneously-optimal sampling locations (batch acquisition). This would in principle require maximizing a joint acquisition function (as a function of multiple locations), which is a high-dimensional multi-modal problem. However, this can be approximated in a simpler way by sequentially acquiring a batch of points, each conditioned to the previous ones using the Kriging believer method \cite{parallel_1, parallel_2, parallel_3, parallel_kriging_believer_1, parallel_kriging_believer_2, update_equation}. In that method one optimizes the acquisition function, conditions the GP on the emulated mean $\mu$ at the previous maximum (which is a comparatively cheap operation)\footnote{This is because only the kernel matrix changes in this step, while the hyperparameters do not need to be refitted. Indeed, the highest cost of this operation is solely a single kernel matrix decomposition and inversion required for future predictions on this conditioned GP.} 
and recomputes the acquisition function using this conditioned GP. The true posterior can then be evaluated in parallel at these locations. Throughout this paper, we call this procedure \emph{sequential optimization}.

\subsection{Nested sampling}

Nested sampling (NS) \cite{skilling_nested_sampling,multinest_1,multinest_2,multinest_3,Handley:2015fda,Handley:2015vkr,Higson_2018,Speagle_2020} is a family of Monte Carlo sampling algorithms and, simultaneously, an integrator for probability density functions (or positive functions in general). 
It is based on the idea that the marginal likelihood computation can be substituted by a one-dimensional integration:
\begin{equation}
    \int L(\vec{x})\pi(\vec{x}) \diff \vec{x} = \int_0^1 \mathfrak{L}(X) \diff X
\,,
\end{equation}
where $\mathfrak{L}(X)$ is defined as the inverse of the cumulant prior mass containing  only likelihood values greater than a given threshold $\lambda$:
\begin{equation}
    X(\lambda) = \int_{L(\vec{x})>\lambda} \pi(\vec{x})\diff \vec{x}
\,.
\end{equation}
The function $\mathfrak{L}(X)$ is then sampled in increasing order by narrowing (nested) regions that contain only posterior values greater than this threshold. This is performed by tracking a set of \emph{live} points, and sequentially discarding the one with the lowest likelihood value and substituting it for a newly-sampled one. The discarded point is weighed correspondingly to the estimated posterior volume contained within the prior shell defined between the likelihood value of the discarded point and the one of the next lowest-likelihood live point. Due to the nature of NS as an integration, Monte Carlo samples from NS produce a better representation of the dynamic range of the distribution than other MC samplers. A review of possible implementations and application to physical sciences can be found in \cite{Ashton:2022grj}.

In this paper, we will use the publicly available \textsc{PolyChord} code \cite{Handley:2015fda,Handley:2015vkr}, in particular the \textsc{PolyChordLite} python wrapper available at \url{https://github.com/PolyChord/PolyChordLite}. The advantage of using this implementation: the code is well known to allow for massively parallel exploration of the desired function (see \cite{Handley:2015vkr} for the weak and strong scaling), due to the use of slice sampling to sample from constrained likelihood contours it scales mildly with dimensionality, and due to its cluster identification algorithm it also very good at identifying global maxima even when multiple local maxima are present (see e.g. Rastrigin example in \cite{Handley:2015vkr}).

\section{Method}\label{sec:method}

\subsection{Monte Carlo sampling}

The basis of our method is substituting the sequential optimisation of the acquisition function using Kriging believer by the exploitation of a Monte Carlo sample of the mean of the GP. Individual samples are ranked according to their acquisition function, as explained below, in a way that reproduces a conditioned ranking similar to what would be obtained via sequential optimization.

Since the target of the acquisition procedure is the optimisation of the aquisition function, it might seem most logical to generate the MC sample directly from it. Nevertheless, there are a number of convincing arguments in favor of sampling on the mean of the GP instead:
\begin{description}
    \item[Speed:] Predicting the GP mean and standard deviation at multiple points simultaneously is much faster due to the possible use of vectorized matrix multiplication routines, but such vectorization is hard to exploit during optimization. While the prediction of the mean is a matrix multiplication of size $(N_\mathrm{new},N_\mathrm{train}) \times N_\mathrm{train}$, the evaluation of the standard deviation requires at least the matrix multiplication of size $(N_\mathrm{train},N_\mathrm{train}) \times (N_\mathrm{train},N_\mathrm{new})$.\footnote{~There is also a trace of a matrix product, requiring additional $N_\mathrm{train} \cdot N_\mathrm{new}$ operations, but this is always subdominant in runtime.} As such, it is often times cheaper to first predict only the mean during the sampling (sequential) and then evaluate the standard deviation. These are then used for the acquisition function computation in a single vectorized call.
    \item[Simplicity:] The acquisition function is often very multi-modal and rather difficult to sample while the mean for a typical well-behaved likelihood is comparatively simple. This reduces the runtime of the MC sampler (sometimes quite drastically) for a given convergence criterion of the MC sample.
    \item[Regions of Interest:] While there almost surely exist regions far away from the mode with large standard deviations and corresponding acquisition function values, it is not always a good idea to actually sample these. This is because the actual posterior mode defines a region of interest where the accuracy of the GP is desired to be high, while other regions are not necessarily important to sample. This becomes especially interesting in moderately-high dimensionality where the volume contained by the mode becomes an ever smaller fraction of the total prior volume. However, we stress that the nested sampling employed in this work typically explores all regions relevant to the acquisition function in our examples -- This region of interest is thus not an entirely strict notion.
    \item[Reusability:] Since the nested sampling run is performed on the mean of the GP, this is effectively giving us a sample of the emulated posterior at this step, useful for inferring marginal quantities (such as credible intervals, means, variances, marginal distributions, etc.).
\end{description}

The use of NS in particular is advantageous with respect to Markov-chain Monte Carlo methods in this particular case: it naturally balances \emph{exploration} and \emph{exploitation}, since it samples the full dynamic range of the target distribution, including its tails, where low-value-but-high-variance optimal locations dwell; it is also almost-embarrasingly parallel up a number of processes similar to the number of \emph{live} points tracked during sampling, and, depending on implementation, has a mild divergence with dimensionality (true for \textsc{PolyChord}).

After all samples have been drawn, we compute the acquisition function at these locations simultaneously and use the result to create a batch of new active sampling location proposals.\footnote{The authors of \cite{Pellejero-Ibanez:2019enw} also perform an MC of the mean GP, but do not take care of conditioning when selecting optimal candidates, as we do in the next section.}

\subsection{Ranked acquisition pool}

Instead of the sequential optimisation approach discussed in \Cref{sec:acq}, we develop an algorithm to rank the MC samples according to their acquisition function value conditioned to the rest of the candidates, using Kriging believer, until an optimal batch of candidates is found. We call this approach a ranked acquisition pool (RAP). To rank a set of points, we start with the sample with the highest unconditioned acquisition as our accepted starting point. From there, we condition the GP to the already accepted samples, and rank all other points according to their acquisition function value conditioned to those accepted samples (i.e. compute the acquisition using the uncertainty of the GP conditioned to the accepted samples). We include an empty slot at the bottom of the pool for temporary sorting. Any sample in that slot will be eventually discarded. Importantly, for any acquisition function monotonic in the GP uncertainty the conditioning can only lower the acquisition value of a point.

\begin{figure}
    \centering
    \includegraphics[width=0.3\textwidth]{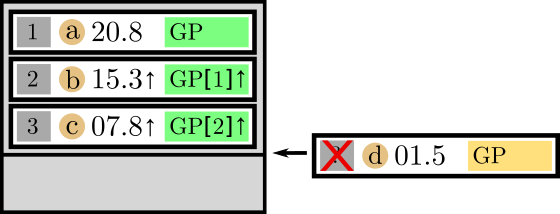}
    \vspace*{1em}
    \hrule
    \vspace*{1em}
    \includegraphics[width=\textwidth]{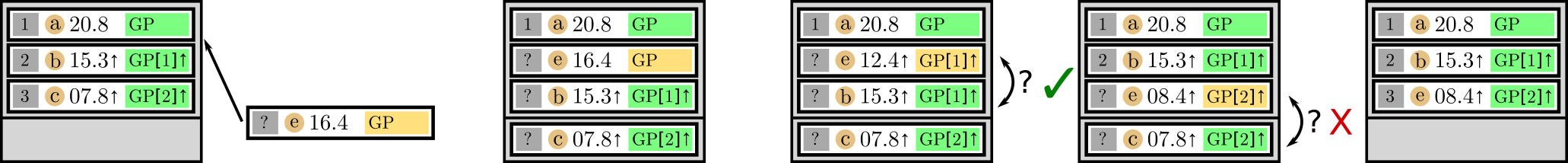}
    \vspace*{1em}
    \hrule
    \vspace*{1em}
    \includegraphics[width=0.7\textwidth]{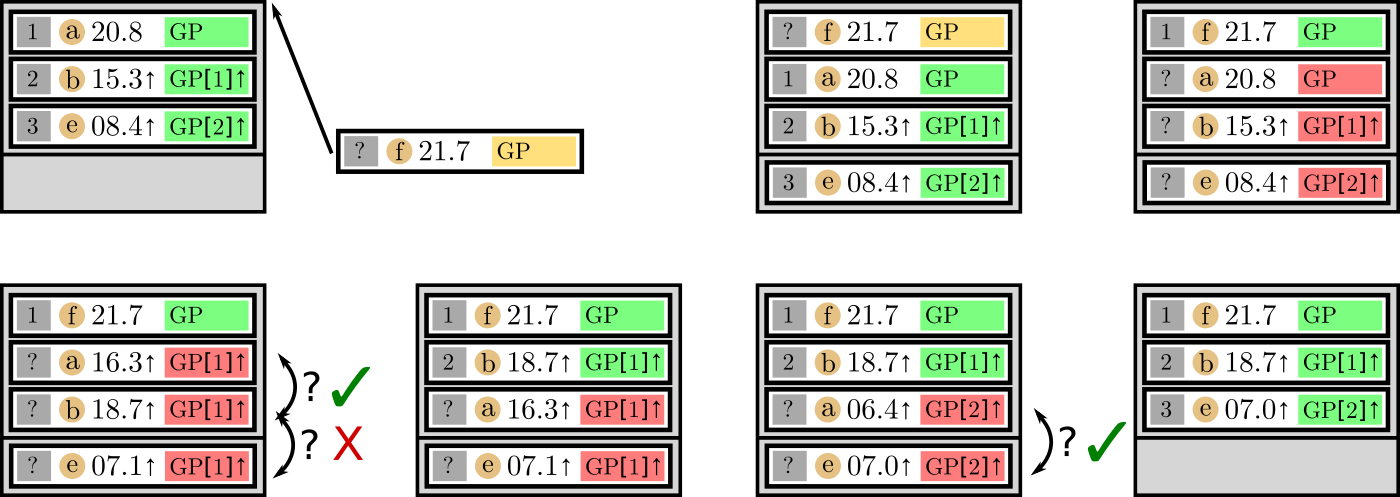}
    \caption{Three insertion cases in a ranked pool of size 3, with one empty slot below it. Each rectangular box represents a single sample. The number in the grey box represents current known rank of the sample (? means un-ranked), the letter in the orange circles is an identifier of the points, the number next to it is the current (conditioned) acquisition function, and the green/orange/red box at the end shows if a point is conditioned to a given rank and those above it (number in angular brackets) and the status of the acquisiton value (green=up-to-date, orange=newly inserted and possibly in need of conditioning, red=invalidated by insertion at higher rank). The three cases are described in the main text.}
    \label{fig:ranked_insertion}
\end{figure}

We separate the algorithm of proposing a new sample into three main steps, and make use of \Cref{fig:ranked_insertion} to show examples for each (description in italics at the end of a step).
\begin{enumerate}
    \item Initial rejection: A sample is only added if its unconditioned acquisition function is larger than the lowest conditioned acquisition function.
    \textit{The sample $d$ is rejected from the acquisition pool since its unconditioned acquisition function is smaller than those of samples $a,b,c$ already present in the pool.}
    \item Insertion and conditioning: If a sample is not rejected, it is initially inserted at the rank corresponding to its unconditioned acquisition function. If it isn't inserted at the top, it has to subsequently be conditioned to all the points above it (which typically decreases its acquisition function). If it is now lower than the next rank, it is inserted and re-conditioned there. This process is repeated until it is higher than the next rank (goes to step 3), or at the bottom of the pool and thus rejected.
    \textit{Sample $e$ is proposed to the pool, and in its unconditioned state ranks in the second position. However, after conditioning it to the first point, it performs worse than sample b and is pushed one rank down. It is then conditioned to the two points above it. This time, it performs better than sample $c$ and thus is inserted into its current position. Since its current position is the last position of the pool no resorting is necessary.}
    \item Resorting: If a sample has been inserted at any rank but the lowest, all the other ranks below are now conditioned to the wrong samples, and need to be re-conditioned and correspondingly re-ranked. This happens in an iterative fashion, where all samples in the current pool compete for the next highest position under the inserted sample (using the same conditioned GP), and the highest conditional acquisition sample is inserted there. Then the process repeats until all the slots have been filled.
    \textit{The element $f$ is added to the pool. Its unconditioned acquisition function places it at the top, and it does not need to be conditioned. This invalidates all other ranks, necessitating a full re-sorting of the pool. Next, all of ($a$,$b$,$e$) compete for the second slot by computing the acquisition function value conditioned to the first rank (here sample $b$ wins). Samples $a$ and $e$ now compete for the third slot by computing their acquisition value when conditioned to ranks 1 and 2 simultaneously (sample $e$ wins).}
\end{enumerate}

In order to speed up especially the computations of the conditional acquisition function, the ranked pool works with a cached model of the GP regressor instances, in order to quickly compute acquisition function values conditioned to a certain rank (and those above it). A technical description of the algorithm can be found in Algorithm~\ref{alg:ranked_pool}.

\begin{algorithm}
\begin{algorithmic}[1]
    \Text{Known:}{ Samples $X_1 ... X_N$ with stored conditional acquisitions $a[0]...a[N]$}
    \Require New sample $X$, a($X$)
    \Statex \Comment{Rejection check}
    \State $i \gets N$
    \If{a(X) > a[N]}
        \State reject($X$)
    \EndIf
    \Statex \Comment{Finding the correct insertion position}
    \While{i>0}
        \State c = $a(X) | (i-1), ..., 1$
        \If{c > a[i-1]}
            \State $i \gets i-1$
        \Else{}
            \State insert($X$,$i$)
        \EndIf
    \EndWhile
    \Statex \Comment{Resorting + rebuilding the cache:}
    \While{i < N-1}
        \For{j \textbf{in range}(i+1,N)}
            \State compute $c_j = a(X_j) | i, ..., 1$\Comment{Update conditioned acquisitions}
        \EndFor
        \State m = argmax[$c_{i+1},...,{c_n}$]
        \State swap($X_{i+1}$,$X_m$)
        \State $a[i+1] \gets c_{m}$
    \EndWhile
\end{algorithmic}
\caption{The ranked pool updating routine in pythonic pseudo-code.\label{alg:ranked_pool}}
\end{algorithm}

By giving up maximization in favour of sampling, our candidates are not the true optima of information gain, but they will be close enough to them. It is more important to get a batch of near-optimal candidates at the same time than getting just a few perfect ones.

\section{Results}\label{sec:results}

\begin{figure}
    \centering
    \includegraphics[width=\textwidth]{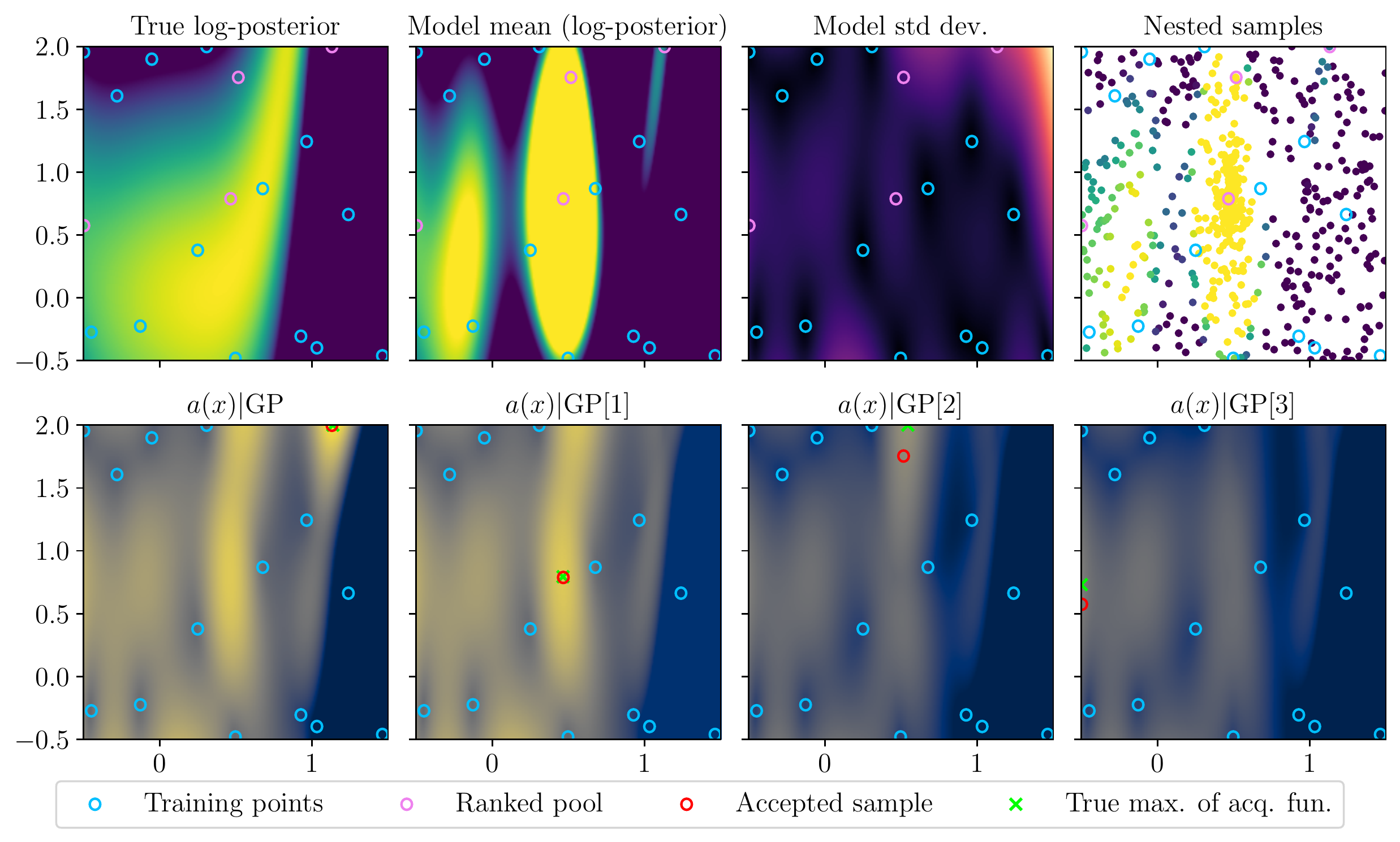}
    \caption{Acquisition procedure with a ranked pool of size $N=4$. The top row shows from left to right: The true function to be emulated, the current GP mean prediction, it's standard deviation, the nested samples (dead points) from \textsc{PolyChord}. The bottom row shows the acquisition function for the unconditioned GP on the left, and for the conditioned GPs in the three right panels (each conditioned to all samples added to its left). Blue circles are current training samples, pink circles are samples that have been accepted into the ranked pool (top), and red circles are each respective optimal sample for the conditioned GP (bottom). Note that this example is very early in the active sampling so the mode has not been well mapped. Nevertheless it is visible that even with very few samples the locations of the nested samples still cover the regions of high acquisition function well.}
    \label{fig:acquisition_2d}
\end{figure}

The combination of the nested sampling approach with the ranked acquisition pool is implemented as NORA (\textbf{N}ested sampling \textbf{O}ptimization for \textbf{R}anked \textbf{A}cquistion), based on the GP treatment from \cite{scikit-learn,Gammal:2022eob} (as well as useful functionality from \cite{2020SciPy-NMeth,harris2020array}). A demonstration of the acquisition procedure in NORA can be found in \cref{fig:acquisition_2d}.

We tested NORA on a number of synthetic likelihoods to demonstrate both the accuracy and the highly parallel nature of our approach. The likelihoods for accuracy tests include a curved degeneracy, a ring, and the multi-modal Himmelblau function. Further discussion of these synthetic examples can be found in \cref{app:sec:test_examples},  while real-world applications to cosmological data can be found in \cref{app:sec:cosmo_examples}.

\begin{figure}
    \centering
    \includegraphics[width=0.45\textwidth]{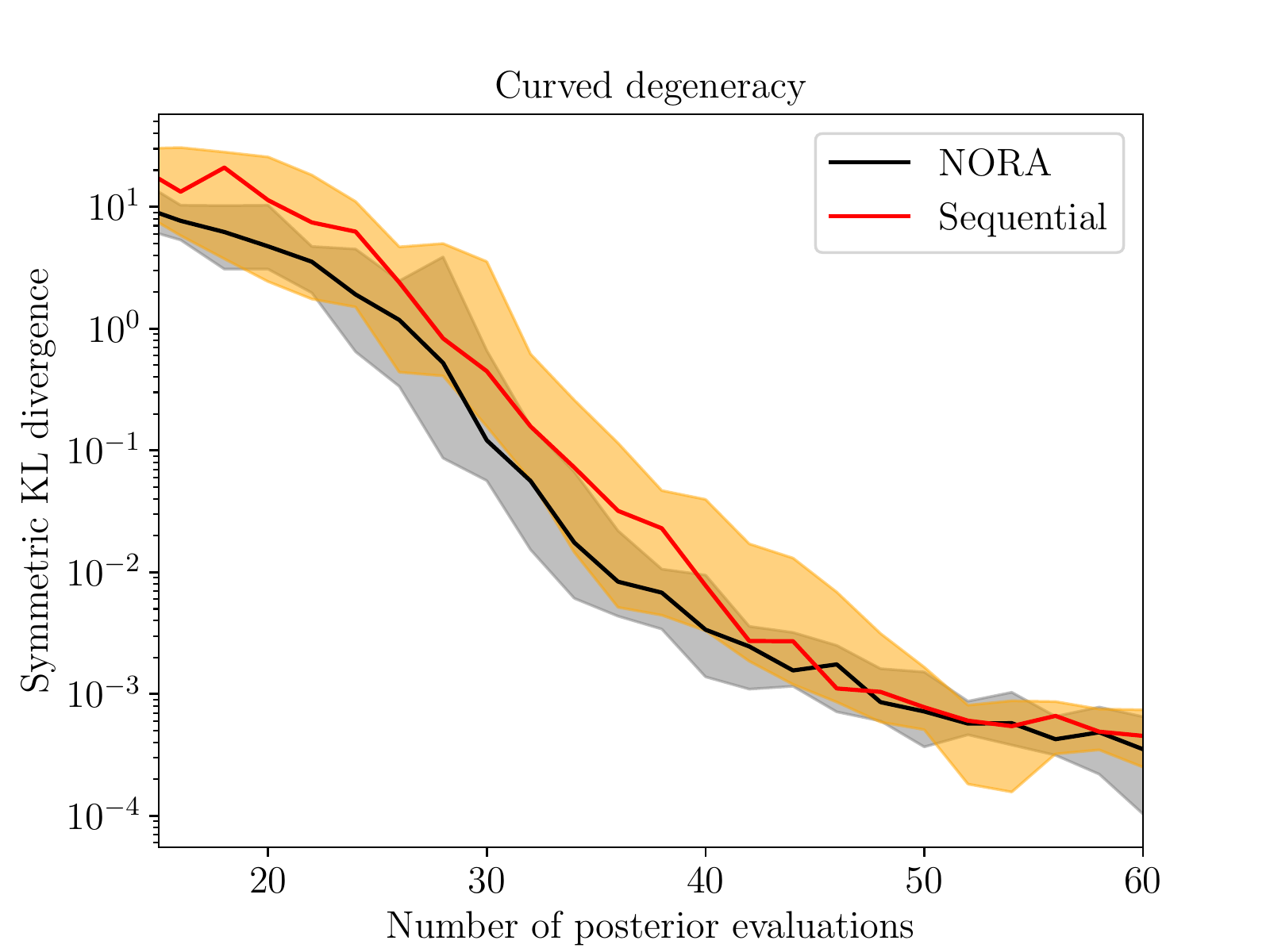}
    \includegraphics[width=0.45\textwidth]{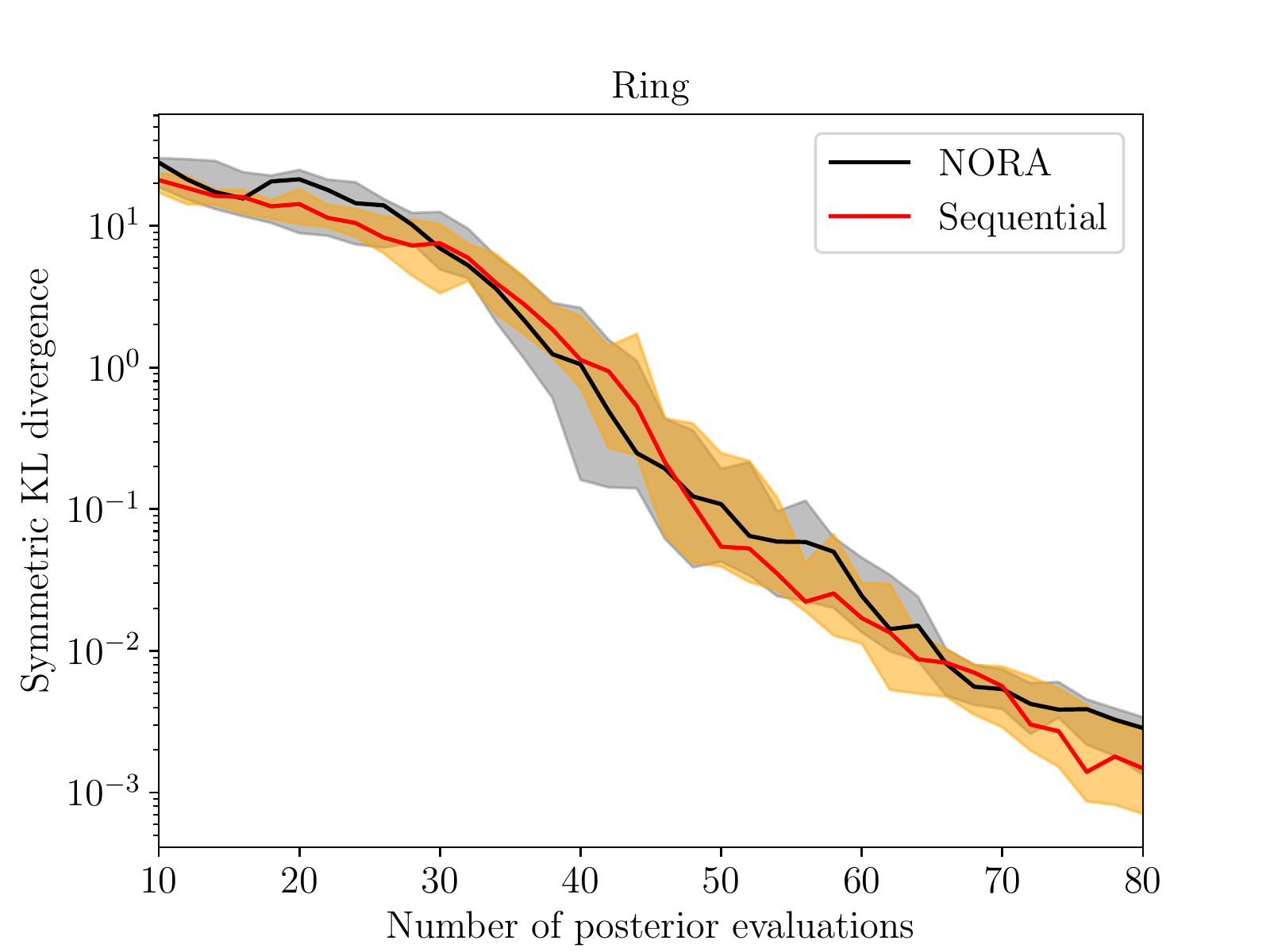}
    \caption{Comparison of the efficiency and accuracy of the acquisition procedure between the naive sequential optimization approach and the NORA approach. We show the agreement between the emulated and the true posterior (specified by the symmetric KL divergence) as a function of the number of samples (posterior evaluations). The solid line is the median, and the shaded region is the 25\% to the 75\% quantiles of 20 realizations. In this case NORA shows similar performance to the sequential algorithm.}
    \label{fig:acq_nora_comparison_curv_ring}
\end{figure}

\begin{figure}
    \centering
    \includegraphics[width=0.45\textwidth]{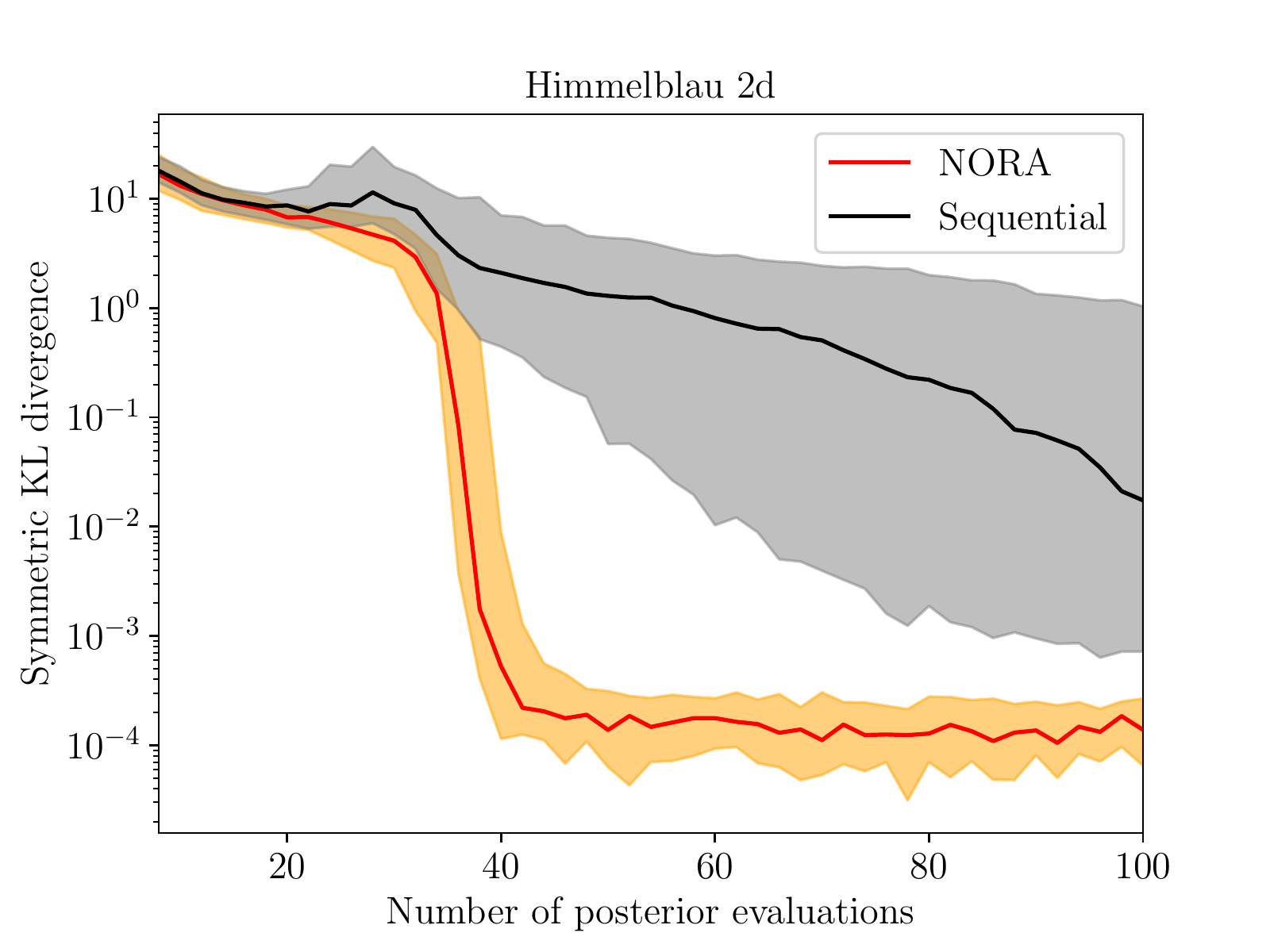}
    \includegraphics[width=0.45\textwidth]{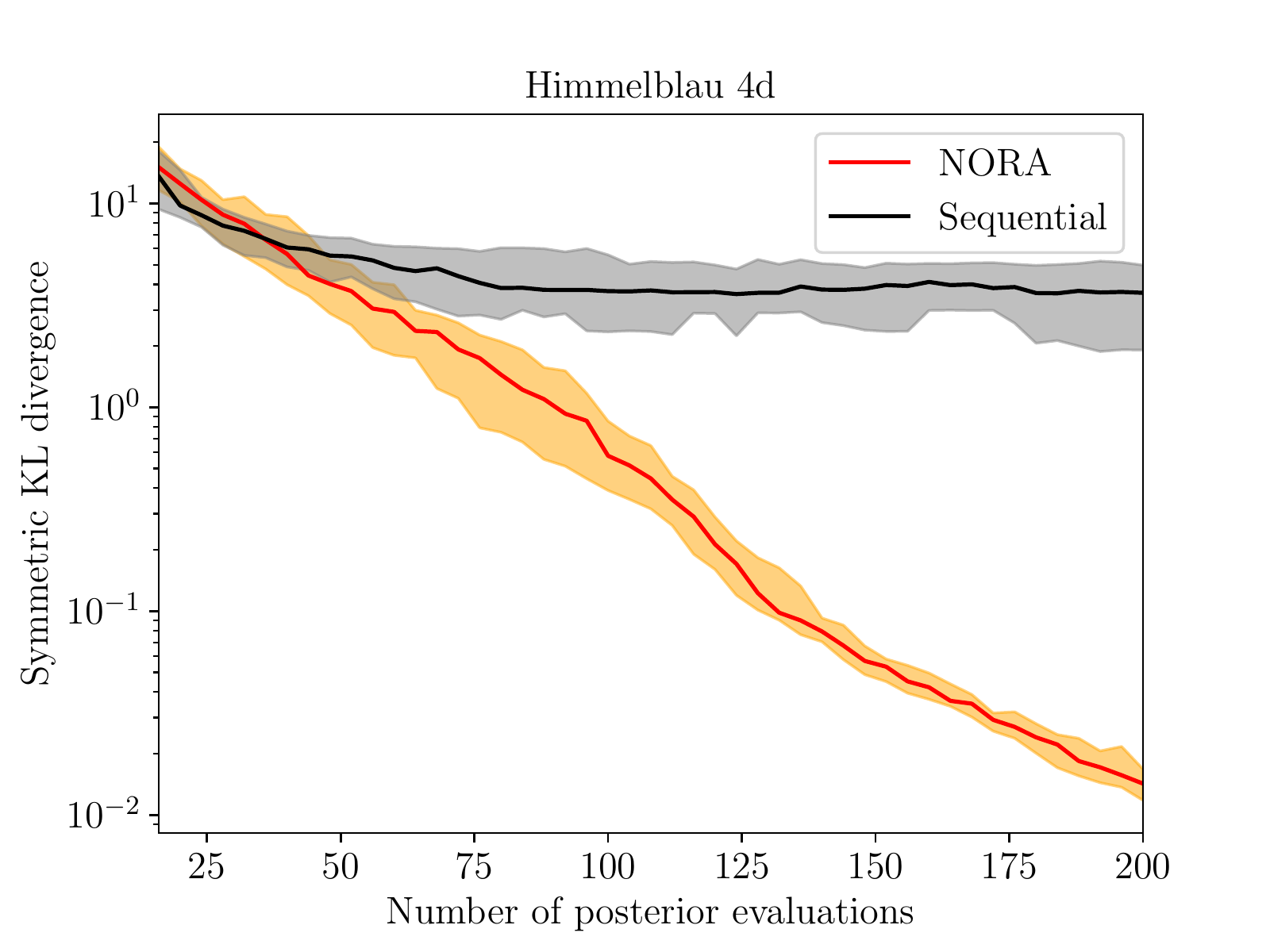}
    \caption{Same as \Cref{fig:acq_nora_comparison_curv_ring} for the Himmelblau function (left) and a four-dimensional extension with 4 modes in two of the dimensions (right). In the other two dimensions it is flat. In these multi-modal cases the NORA algorithm is far more efficient than the sequential sampling algorithm.}
    \label{fig:acq_nora_himmelblau_2d}
\end{figure}

\begin{figure}
    \centering
    \includegraphics[width=0.4\textwidth]{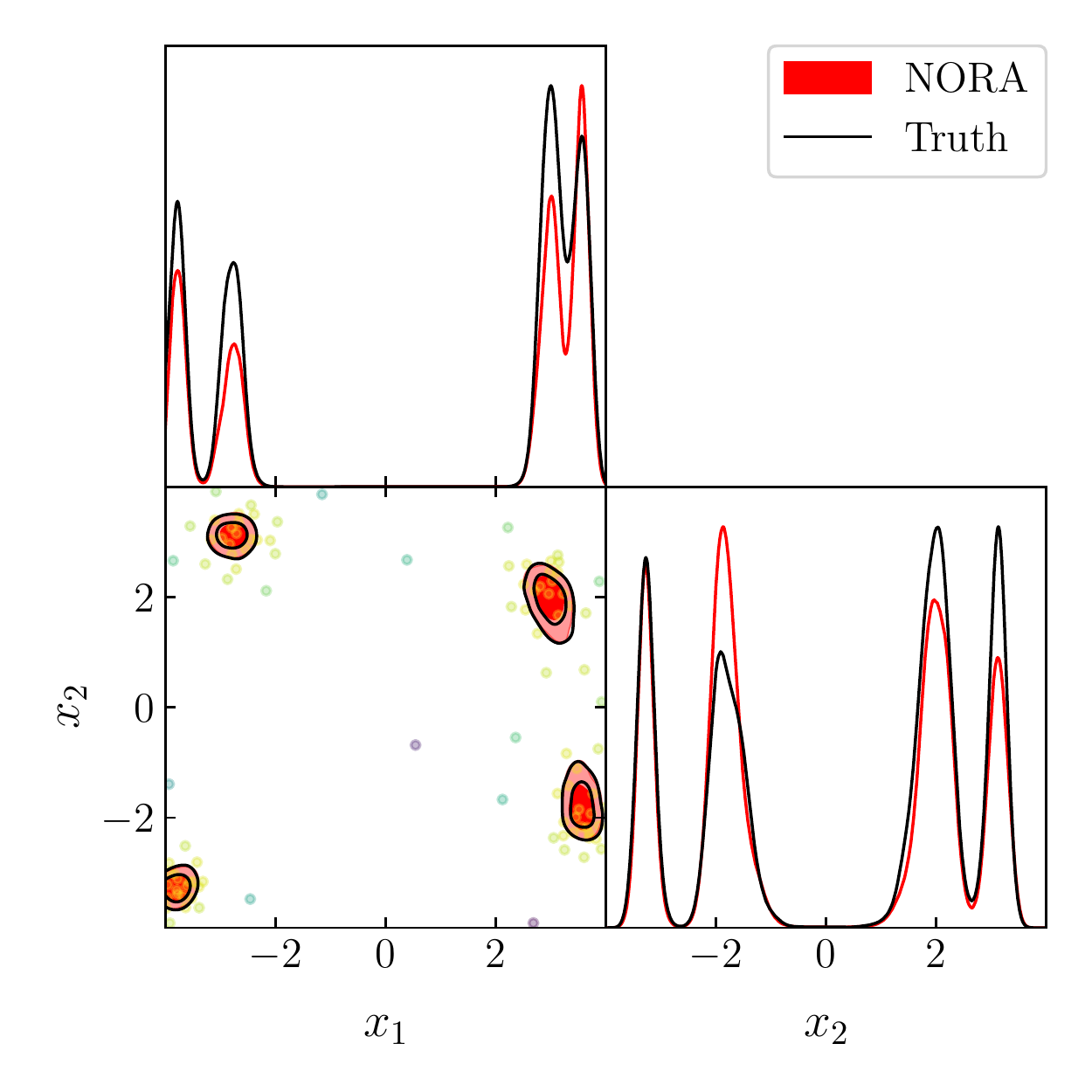}
    \includegraphics[width=0.4\textwidth]{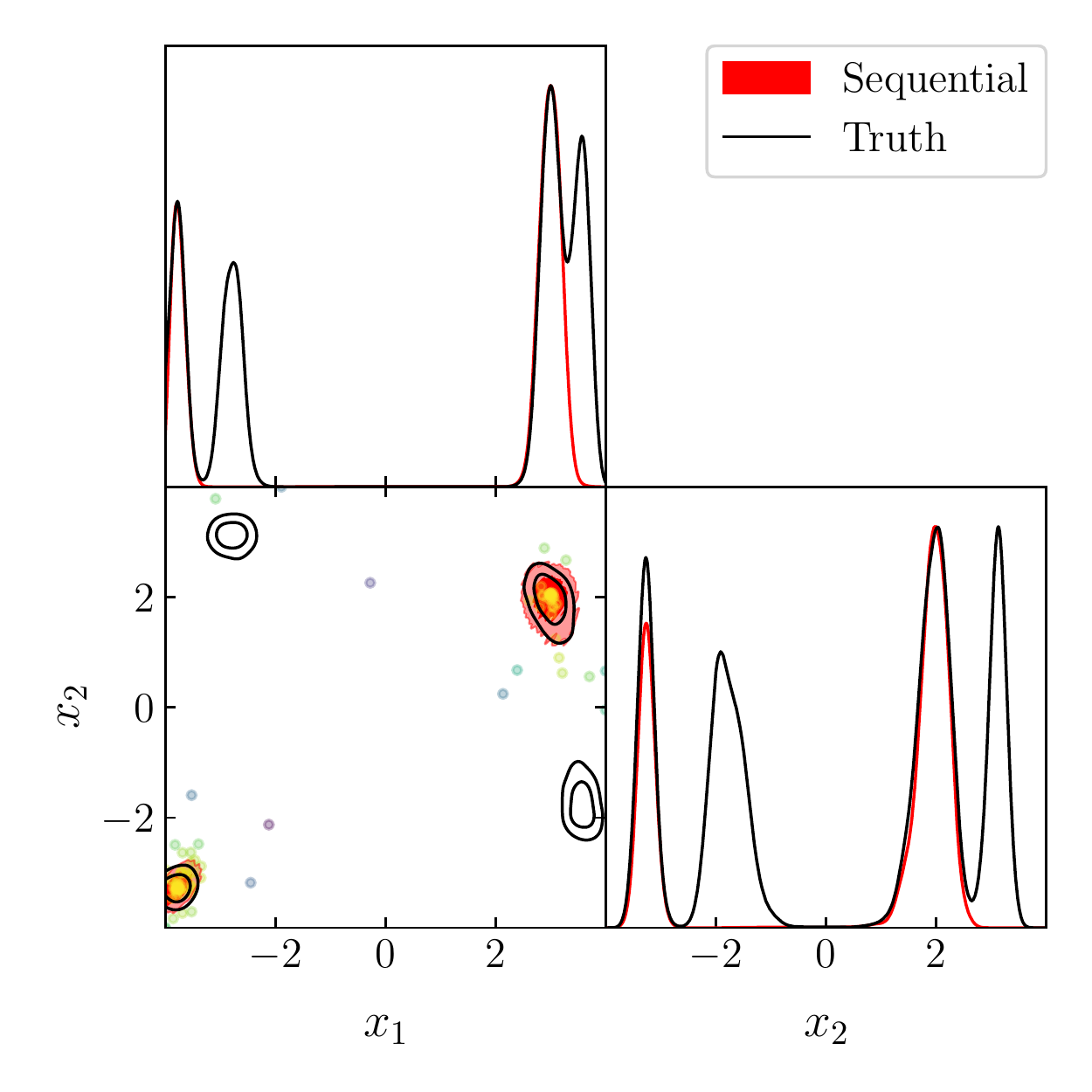}
    \caption{Example of a failure case of the naive sequential optimization compared to the same cases treated with NORA, which due to its nested sampling correctly identifies all modes. The sampling locations are shown with colour denoting how late they were sampled (yellower$=$later). It is clearly visible that the sequential optimization is sampling more aggressively towards the top of the mode and showing less explorative behaviour. Both are allowed 100 posterior evaluations.}
    \label{fig:2d_himmelblau_contours}
\end{figure}

The curved degeneracy (see also \cite{curved_degeneracy,Gammal:2022eob}) has a tight ridge in the $x_2 \approx 4 x_1^4$ direction, and its log-likelihood is
\begin{equation}
    \log L(x_1, x_2) = -(10\cdot(0.45-x_1))^2/4 - (20\cdot(x_2/4-x_1^4))^2 \ .
\end{equation}
The log-likelihood for the ring example is instead
\begin{equation}
    \log L(x_1, x_2) = -\frac{1}{2}\left[\frac{(\sqrt{x_1^2+x_2^2}-\mu)^2}{\sigma}+\log(2\pi\sigma^2)\right]~,
\end{equation}
where $\mu=1$ and $\sigma=0.05$ in our example. We show in  \Cref{fig:acq_nora_comparison_curv_ring} that in both of these cases the accuracy and efficiency is very comparable to the sequential method (while much more parallelizable, see below). Both reach about the same level of agreement between emulation and the true function (as captured by their symmetric KL divergence, which is further explain in \cref{app:sec:kl}). We also investigate a multi-modal example like the Himmelblau function with log-likelihood
\begin{equation}
    \log L(x_1,x_2)=-\frac{1}{2}\left[100\cdot(x_1^2-x_2-11)^2+(x_1+x_2^2-7)^2\right]
\end{equation}
We furthermore construct a four-dimensional version of this function which retains the four maxima in two dimensions but is constant along the other two dimensions. This combines the multimodality with the problem of correctly mapping and exploring the flat dimensions.
We show the results in \Cref{fig:acq_nora_himmelblau_2d}. Since in this case the nested sampling has a far higher chance of quickly discovering a mode of the function far from the already known ones, the NORA approach is much more efficient than the Sequential optimization approach in this case (we show examples of explicit modeling for 100 posterior evaluations in \Cref{fig:2d_himmelblau_contours}).

In order to assess that gains in modelling do not come at the cost of overhead in the acquisition step, we have performed a number of tests in Gaussian likelihoods at different dimensionalities. The comparison with the costs of acquisition with sequential optimization and NORA, as well as the scaling with parallelization is shown in \Cref{tab:scaling_nora}. We see that the overhead of NORA is comparable to that of sequential optimization for the same number of MPI processes. However, sequential optimization will only profit from parallelization up to the number of restarts of the optimizer while nested sampling will parallelize virtually infinitely (up to the large number of live points).


\begin{table}
    \centering
    \begin{tabular}{c | cc|cc|cc}
          &\multicolumn{2}{c}{\bf $d=2$} &
         \multicolumn{2}{c}{\bf $d=4$} &
         \multicolumn{2}{c}{\bf $d=8$} \\ \hline \rule{0pt}{2.4ex} 
         SeqOpt & [2] & [2.4, \textbf{2.55}, 2.7] &  [4] & [11.2, \textbf{12.1}, 12.7] &  [8] & [177, \textbf{183}, 191] \\ \hline 
        NORA &\rule{0pt}{2.4ex}  [2] & [3.67, \textbf{3.84}, 4.34] &\multicolumn{2}{c}{---}& \multicolumn{2}{c}{---}\\
         NORA & [4] & [1.43, \textbf{1.54}, 1.75] &  [4] & [9.3, \textbf{9.9}, 10.6] & \multicolumn{2}{c}{---} \\
         NORA & [8] & [0.96, \textbf{1.05}, 1.13] &   [8] & [6.29 \textbf{6.58} 6.90] &  [8] & [147, \textbf{160}, 220] \\
          NORA &[16] & [0.30, \textbf{0.33}, 0.35] &  [16] & [4.52, \textbf{4.77}, 5.16] &  [16] & [122, \textbf{129}, 171] \\
    \end{tabular}
    \caption{Comparison of wall-clock runtimes for the acquisition step between NORA and sequential optimization (dubbed "SeqOpt", with 5$\,\cdot\, d$ restarts of the optimizer). We add in angular brackets the number of MPI processes. In each dimensionality we show the [25, \textbf{50}, 75] percent quantiles. We run 50 runs in 2- and 4 dimensions, and 20 runs in 8 dimensions, with respective truth evaluation budgets 20, 60 and 400. Convergence in terms of symmetric KL divergence is similar in all cases and of magnitude $\mathcal{O}(0.01)$. We additionally allow multi-threading (useful e.g. for BLAS \cite{blackford2002updated} matrix operations) for each MPI process up to a total of 32 cores.}
    \label{tab:scaling_nora}
\end{table}

\section{Conclusion}\label{sec:conclusion}
Sequential optimization for active learning is facing a variety of challenges, such as difficult parallelization, and a lack of robustness to getting stuck in local maxima, thus requiring many restarts of the optimizer in high dimensions to properly explore the target inference space. To overcome these challenges we propose a new algorithm, called NORA, that substitutes the sequential optimization of the acquisition function by combining Monte Carlo exploration of the GP's mean using Nested Sampling, and ranking of the Monte Carlo samples according to their conditional acquisition function values, to generate a nearly optimal batch of sampling locations. These two steps can be performed in a nearly perfectly-parallelizable way, and the same Monte Carlo sample can be reused in consecutive iterations for lowering computational costs.


We apply NORA to a number of synthetic Bayesian inference problems to assess its performance, and compare it to a reasonably good implementation of sequential optimisation of the acquisition function.

We find that NORA and sequential optimization perform equally well at comparable computational costs for simple unimodal likelihoods for $d<10$, and for highly non-Gaussian likelihoods in small dimensionalities. NORA greatly outperforms sequential optimization for multi-modal likelihoods, due to the more exploratory approach to acquisition, despite producing less precise acquisition batches than sequential optimization.


The limitations of the NORA algorithm are similar to those of other approaches to Bayesian inference based on surrogate GP models: Their strong divergence with dimensionality due to the increasingly large number of training points needed for good posterior modelling, and the $\mathcal{O}(n^3)$ scaling when fitting of the hyperparameters of the GP (see also \cite{2015ITPAM..38..252A}). Furthermore, one particular shortcoming of NORA compared to sequential optimization is that due to its less aggressive acquisition it will converge later in simple problems, e.g.\ Gaussian likelihoods. Our methodology also does not address the problem of stochastic likelihood evaluations (see \cite{NEURIPS2020_5d409541}).

\begin{ack}
    J.~T.~acknowledges support from the STARS@UNIPD2021 project \textit{GWCross}.
    N.~S.~acknowledges support from the Maria de Maetzu fellowship grant: CEX2019-000918-M, financiado por MCIN/AEI/10.13039/501100011033. J.~E. acknowledges support by the ROMFORSK grant project no.~302640.
\end{ack}

\include{supplementary}

\bibliography{biblio}

\end{document}

%% file: supplementary.tex

\appendix

\section{Description of the surrogate model}
In this section we describe details of the GP surrogate model (see also \cite{Gammal:2022eob} for a detailed description).

\subsection{Choice of kernel function}\label{ssec:kernel}

On top of the choice of the kernel function itself, as defined in \Cref{eq:kernel_multi}, some knowledge of the target function is also incorporated in the priors for the hyperparameters. Our assumption is that the length scales should be of an order of magnitude close to that of the posterior modes, while the latter would be of an order of magnitude not much smaller than that of the prior ranges for the parameters of the posterior. We express this belief by setting the prior of the length scales to being uniform between $0.01$ and $1$ in units of the prior length in each direction. This condition assumes that the size of the mode is larger than about $1/100$th of the prior width in each dimension, which we find reasonably permissive. The prior of the output scale $C$ is chosen to be very broad and allows for values between $0.001$ and $10000$. The $d+1$ free hyperparameters $\theta\equiv\{C,l_i\}$ are then chosen such that they maximize 
\begin{align}\label{marginal_gp_likelihood}
-\log p(\vec{y}|\vec{X},\theta) = \frac{1}{2}\vec{y}^T(\vec{k}(\vec{X},\vec{X})+\sigma_n^2 \vec{I})^{-1}\vec{y} + \frac{1}{2} \log|\vec{k}(\vec{X},\vec{X})+\sigma_n^2 \vec{I}| - \frac{N_s}{2}\log 2\pi~.
\end{align}
where $\sigma_n$ is a small noise parameter that typically improves numerical stability of the matrix inversion.
\subsection{Parameter space transformations}\label{ssec:transformations}

To ensure numerical stability we use a number of transformations during the modelling with the GP:

Firstly, we sample the log-posterior distribution to reduce the scale of the function that the GP interpolates. Furthermore, the characteristic length scale of isotropic kernels tends to be larger when sampling the log-posterior, which implies that the GP surrogate generalizes better to distant parts of the function, making the GP more predictive.

In addition, at every iteration of the algorithm, we \emph{internally} re-scale the modeled function using the mean and standard deviation of the current samples set as
\begin{align}
    \log \widetilde{p}(\vec{X}) = \frac{\log p(\vec{X})-\overline{\vec{y}}}{s_{\vec{y}}}\, ,
\end{align}
where $\overline{\vec{y}}$ and $s_{\vec{y}}$ are the sample mean and standard deviation respectively. This re-scaling acts like a non-zero mean function, causing the GP to return to the mean value far away from sampling locations. This in turn encourages exploration when most samples are close to the mode and exploitation when most samples have low posterior values.

As for the space of parameters $\vec{x}$, we transform the samples such that the prior boundary becomes a unit-length hypercube.
This usually leads to comparable correlation length scales of the GP across dimensions, which increases the effectiveness of the limited-memory Broyden-Fletcher-Goldfarb-Shanno (L-BFGS-B) constrained optimizer \cite{L-BFGS-B}, used to optimize the GP hyperparameters.


\section{Some details of the algorithm}

Active-sampling Bayesian inference algorithms based on surrogate models in the literature \cite{NIPS2012_6364d3f0,NIPS2014_e94f63f5,kandasamy:2015,2018arXiv180204782C,10.1162/neco_a_01127,Pellejero-Ibanez:2019enw,Gammal:2022eob,NEURIPS2018_747c1bcc,NEURIPS2020_5d409541,huggins2023pyvbmc} usually follow a fixed procedure: after an initial batch of training samples is either provided or drawn from the prior, the algorithm iterates on a cycle of (1) optimising an acquisition function to obtain candidates for evaluation of the true posterior, (2) evaluation of the true posterior at the proposed locations, (3) refitting of the surrogate model, and (4) convergence checks. In this study we do not concern ourselves with the initial proposal (in our case sampled from the prior) or the convergence checks (in most examples we have fixed budgets of how many true posterior points are sampled), since the focus of this study is on the acquisition step.

As discussed in the main text, our acquisition procedure has two steps: first the mean GP is explored using nested sampling, and, second, the resulting MC samples are ranked according to their conditioned acquisition function value. In this short appendix, we discuss some particularities of these procedures.

\subsection{Scaling of Nested Sampling precision parameters}

The two fundamental parameters of a nested sampler are the number of \emph{live points}, and the fraction of the total posterior mass (evidence) contained in the final set of live points. Additional parameters depend on the particular implementation of NS. In \textsc{PolyChord}, our sampler of choice, the aforementioned parameters are called respectively \texttt{nlive} and \texttt{precision\char`_criterion}. In addition, and among others, \textsc{PolyChord} has two more important parameters: the length of the slice-sampling chains (\texttt{num\char`_repeats}), and the size of the initial prior sample from which the live points are extracted (\texttt{nprior}). It is a natural choice that in the early stages of learning, where limited precision when optimising the acquisition function is enough, this would translate in our procedure into lower precision settings for \textsc{PolyChord}. To reflect this, we scale the number of live points to be proportional to the number of points in the training set (by a factor of 3 by default), with a cap equal to the default precision criterion of \textsc{PolyChord}, which is 25 times the dimensionality of the problem. On the other hand, we have found that the accuracy of our algorithm benefits from more accuracy than the default for the length of slice chains (\texttt{num\char`_repeats}, 5 times the dimensionality instead of 2), whereas the evidence fraction contained in the live points (\texttt{precision\char`_criterion}) can be relaxed with respect to the default by a factor of 5, since we are not interested in an accurate calculation of the model evidence.

\subsection{Byproducts of Nested Sampling}

The nested sampling step produces both a MC sample and a calculation of the evidence of the model. The first one is a useful by-product, which can be used for inference once the run has converged, or to implement a global convergence criterion, such as one based on the calculation of KL divergences between iterations. The value of the evidence is also a useful output, in particular to define a further convergence criterion, but it needs to be taken into account that the resulting NS uncertainty does not include the uncertainty due to the probabilistic nature of the GP, or the uncertainty over the choice of hyperparamenters values, as Bayesian Quadrature approaches do.

\subsubsection{Parallelization}

Nested samplers parallelize effectively up to the number of live points (\texttt{nlive}), since parallel evaluation of the target function increases the chance that at every iteration an acceptable sample will be found at the cost of a single evaluation. Since this number is usually a few tens of times the dimensionality, this step of our algorithm will effectively parallelize linearly with the number of simultanous processes. \textsc{PolyChord} does not do vectorized evaluation of the target function, i.e.\ the target function is always called with a single argument. Hence for this step we prefer to invest CPU cores into separate MPI processes, as opposed to multiple threads.

The ranking step of the algorithm when running NORA in parallel occurs in two steps: first the MC sample is split in as many equal parts as running processes, for evaluation of the GP standard deviation and the acquisition function value, and the individual ranking of each subset into ranked pools with as many points as the desired Kriging believer steps; and later all the ranked pools are combined an re-ranked in a single process. The first of these two steps can be effectively parallelized, but the second one is not parallelizable by definition, and may at most benefit for multi-threading. In most situations, unless the size of the training set is very large, the first step is costlier and thus a larger number of MPI processes is more beneficial than a larger number of threads per process.

Finally, the evaluation step occurs always in parallel when MPI processes are available, but its parallelization is limited by the number of Kriging believer steps we have decided to take. This number must be kept in check because the quality of the batch of proposals decreases when conditioning on increasingly bad information, making our model larger and more computationally expensive. We have found that a number of Kriging believer steps equal to the dimensionality is a good choice in most cases. Highly multimodal posterior can benefit from larger number of Kriging beliver steps, since their acquisition functions have more local maxima, but it would not be wise to go beyond a few times the number of dimensions. Thus, the evaluation step benefits from the number of MPI processes in a limited way, and may be faster if more cores are left available for multi-threading, thus accelerating the evaluation of the true posterior.

The difference between the acquisition step benefiting from a large number of MPI processes, and the evaluation step potentially benefiting more from a large number of threads, makes the choice of the ratio of MPI processes to threads per process dependent on the speed of the evaluation step and the overhead costs, which scale with dimensionality: fast true posteriors in high dimensionality call for larger number of MPI threads, and very slow posterior with an implementation that benefits from multi-threading would call for a larger amount of threads and a smaller amount of MPI processes. In the future, we will look at substituting \textsc{PolyChord} by a nested sampler that can perform vectorized calls to the target function, in order to make multi-threading an overall better choice, beyond the small necessary MPI parallelization for Kriging believer.

\section{Kullback-Leibler Divergences}\label{app:sec:kl}

We define the Kullback-Leibler (KL) divergence of the continuous probability distribution $P$ with respect to $Q$ with probability density functions $p(\vec{x})$ and $q(\vec{x})$ as
\begin{equation}
    D_{\mathrm{KL}}(P||Q) = \int p(\vec{x})\log\left(\frac{p(\vec{x})}{q(\vec{x})}\right)\, d \vec{x}~.
\end{equation}
The KL divergence as defined above more strongly weighs disagreements between the two probability distributions where $p(\vec{x})$ is large. Since we want the approximation to be equally accurate in all regions where either distribution is large, we use a symmetrized version of the divergence (often called Jeffreys divergence). It is defined as
\begin{equation}
    D_{\mathrm{KL}}^{\mathrm{sym}}(P, Q) = \frac{1}{2}\left(D_{\mathrm{KL}}(P||Q) + D_{\mathrm{KL}}(Q||P)\right)~.
\end{equation}
A smaller value means that the two posteriors are in better agreement, and one typically wants $D^{\mathrm{sym}}_{\mathrm{KL}}(P||Q) \ll 1$ for good agreement. The dimensionality consistency of the KL divergence guarantees that a given value for the divergence characterizes similar differences across dimensionalities.

To compute the KL divergence explicitly, one can use the fact that the points in a Monte Carlo sample of $P$ are distributed as $p(\vec{x})d \vec{x}$. One can thus approximate the integral as a sum of the quantity $\log p(\vec{x}_i)-\log q(\vec{x}_i)$ over all points in the MC sample (multiplied by their respective weights/multiplicities). This can be done by evaluating either the real model or the GP emulated posterior for the given points.

There also exists a Gaussian approximation for the KL divergence which is particularly useful when computing the true log-posteriors at each point of the MC sample is computationally undesirable (such as the cosmological examples below). It is defined as
\begin{equation}\label{app:eq:gaussian_kl}
    D_{\mathrm{KL}}(P||Q) \approx \frac{1}{2}\left(\mathrm{tr}\left(\vec{C}_Q^{-1}\vec{C}_P\right)-d+(\mathfrak{m}_Q-\mathfrak{m}_P)^T\vec{C}_Q^{-1}(\mathfrak{m}_Q-\mathfrak{m}_P)+\log\left(\frac{\mathrm{det}\vec{C}_Q}{\mathrm{det}\vec{C}_P}\right)\right)~.
\end{equation}
with $\vec{C}_Q$ and $\vec{C}_P$ being the respective covariance matrices of the two probability distributions, while $\mathfrak{m}_Q$ and $\mathfrak{m}_P$ are the respective means. While the approximation of the individual distribution as multivariate Gaussian is certainly incorrect in non-Gaussian cases, it is typically the case that a good agreement of the Gaussian KL signals a good compatibility of the true KL as well. We always compute the true symmetric KL unless explicitly stated otherwise.

\section{Test functions}\label{app:sec:test_examples}

Here we comment further on the test functions presented in \Cref{sec:results} of the main text.

\begin{figure}
    \centering
    \includegraphics[width=0.6\textwidth]{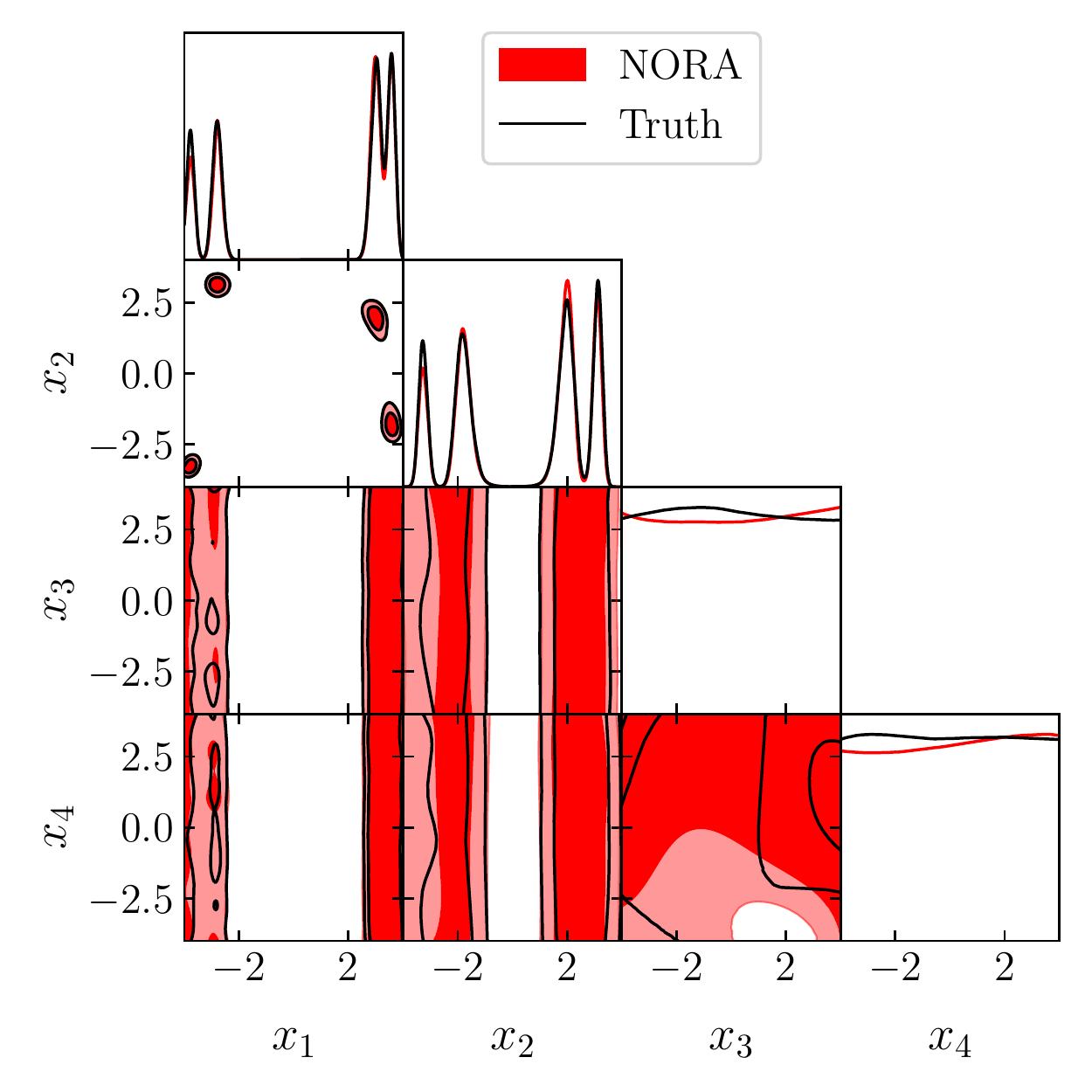}
    \caption{Example of inference on the 4d Himmelblau example. The four modes are in the $x_1$-$x_2$ direction while the other two directions are flat. The example shows NORA sampling with a budget of $200$ posterior evaluations. Both contours are in good agreement with each other.}
    \label{fig:4d_himmelblau_corner}
\end{figure}

\begin{figure}
    \centering
    \includegraphics[width=0.45\textwidth]{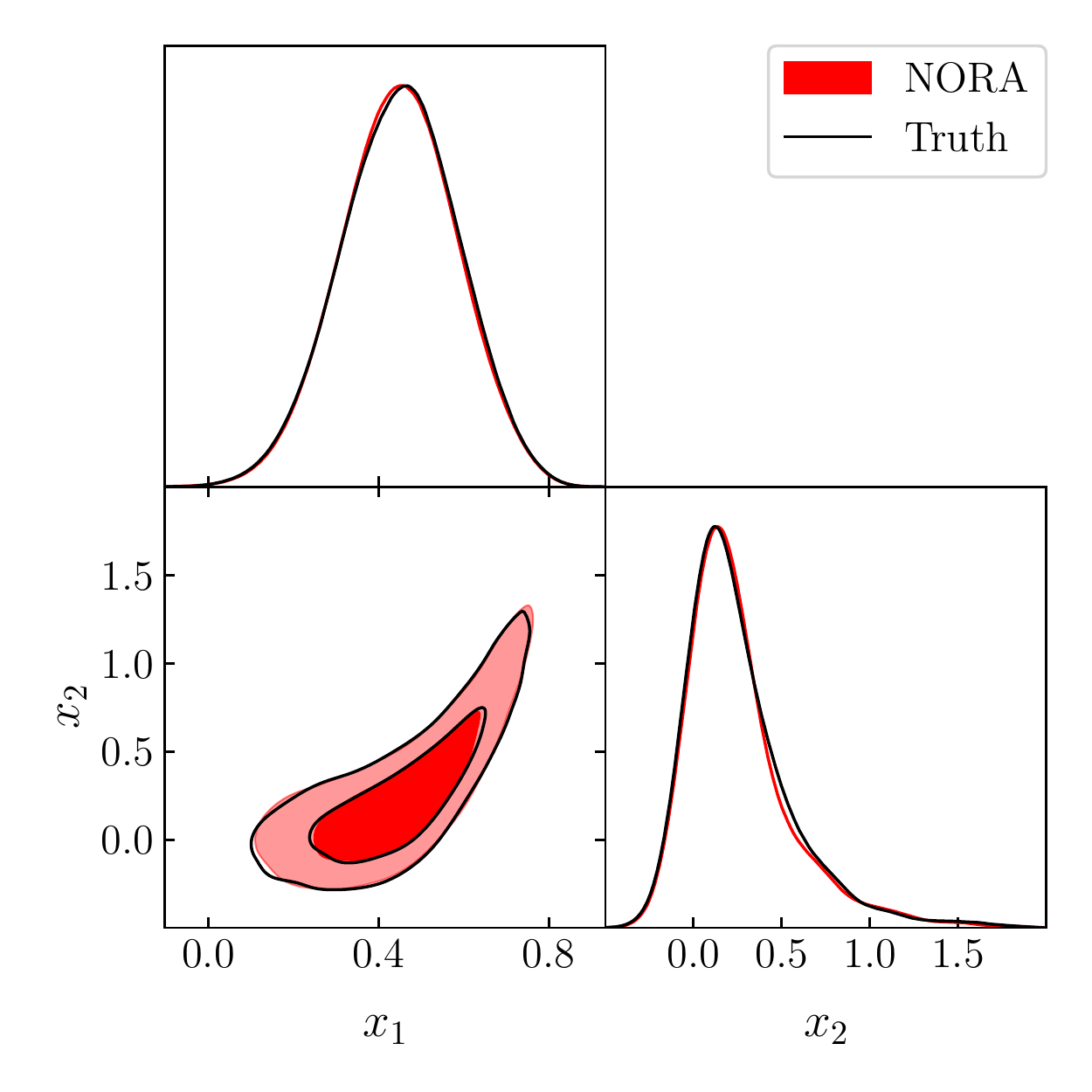}
    \includegraphics[width=0.45\textwidth]{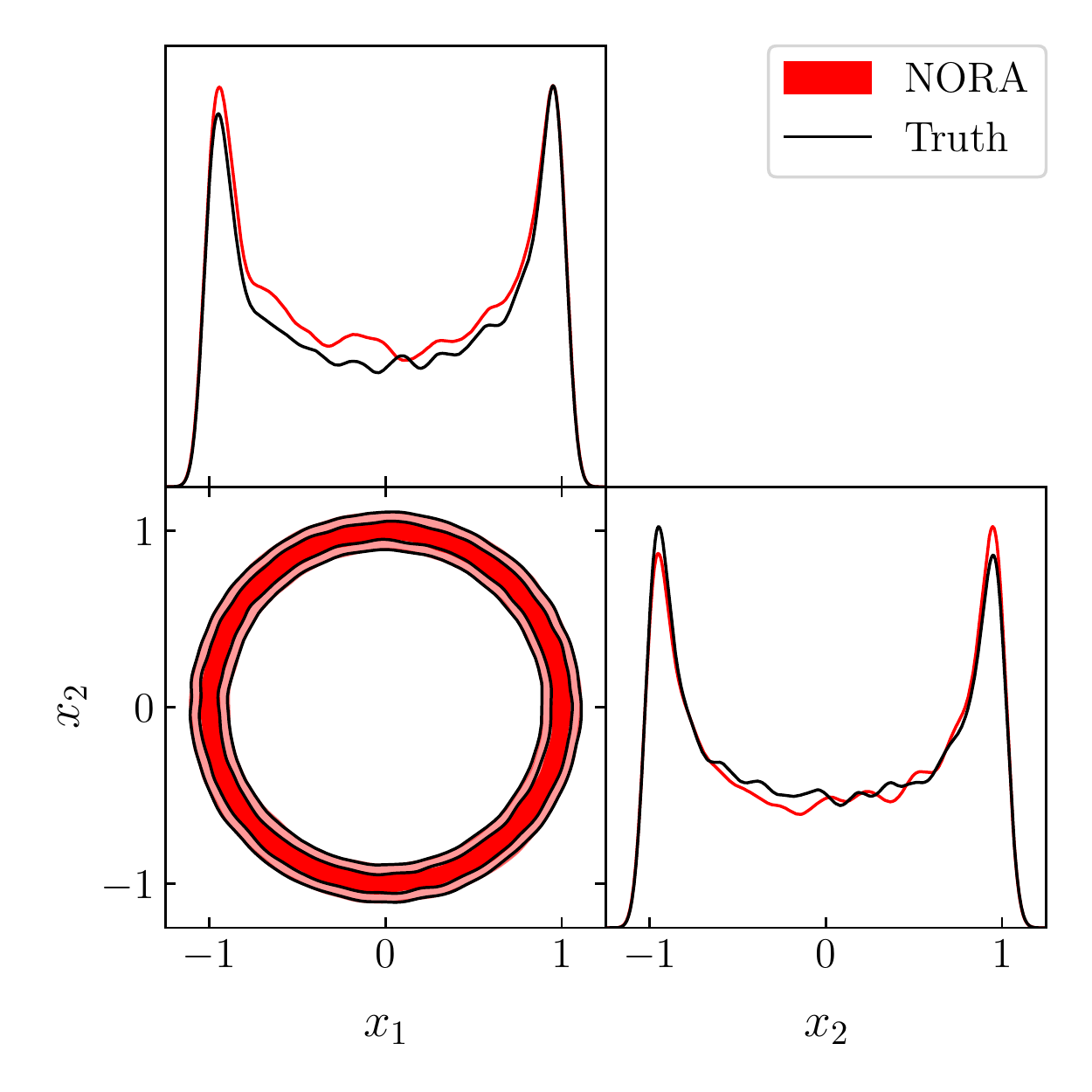}
    \caption{Example of inference on the curved degeneracy example (left) and the ring (right). NORA correctly recovers both contours. Both runs have been performed with a budget of $80$ posterior evaluations.}
    \label{fig:curved_degeneracy_and_ring_corner}
\end{figure}

\Cref{fig:4d_himmelblau_corner} and \Cref{fig:curved_degeneracy_and_ring_corner} show exemplary corner plots of the examples used in \Cref{sec:results}. In all three multi-modal cases presented in that section, NORA correctly recovers the contours. 

\begin{figure}
    \centering
    \includegraphics[width=0.45\textwidth]{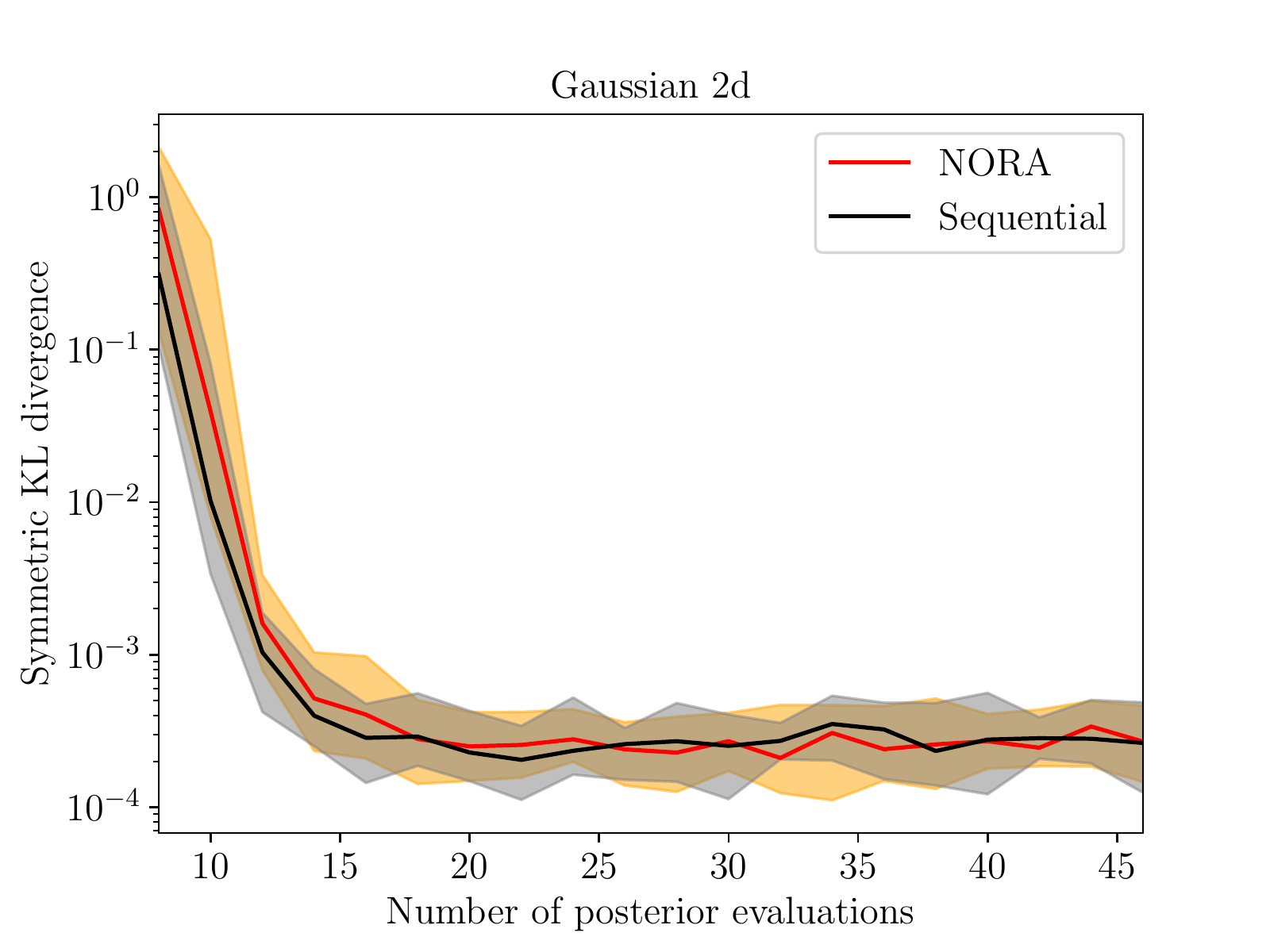}
    \includegraphics[width=0.45\textwidth]{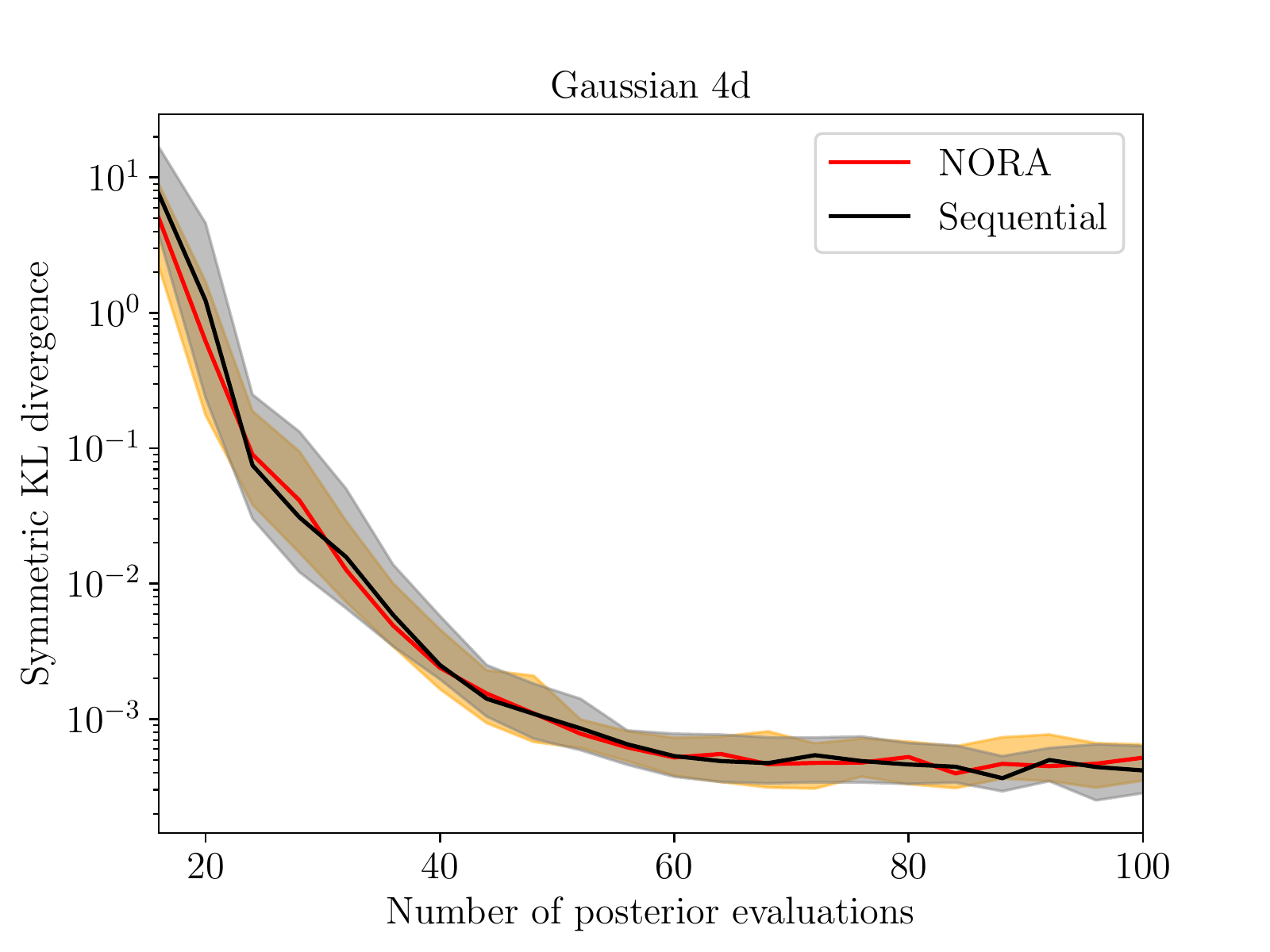}
    \includegraphics[width=0.45\textwidth]{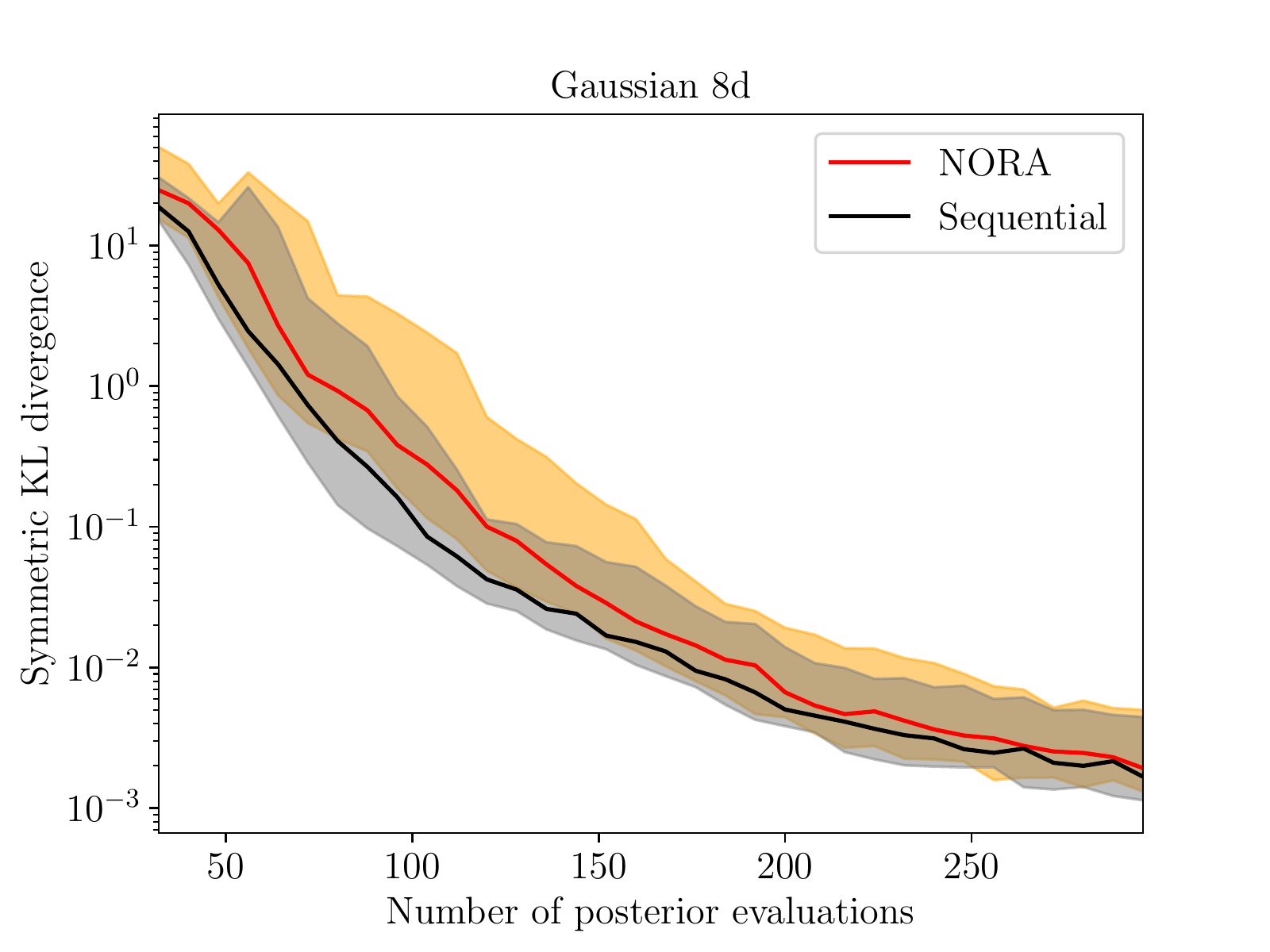}
    \caption{Comparing convergence of NORA vs.\ Sequential optimization for randomly drawn Gaussians in 2, 4, 8 dimensions. NORA and sequential optimization perform nearly equally well for these easy-to-model likelihoods.}
    \label{fig:gaussians_opt_vs_mc}
\end{figure}

In \Cref{sec:results} we also presented a study of the parallelization of the overhead costs of NORA. In this context, in \Cref{fig:gaussians_opt_vs_mc} we show comparisons in convergence between NORA and sequential optimization for Gaussians drawn with random correlations in 2, 4 and 8 dimensions. For these very easy-to-model functions NORA converges as fast as sequential optimization. The slightly slower convergence in $d=8$ is likely due to the somewhat more exploratory behaviour of NORA compared to sequential optimization.

\section{Cosmological examples}\label{app:sec:cosmo_examples}
In order to test the applicability and robustness of the NORA implementation to real-world examples, we also apply it to a number of inference runs commonly used in cosmology. In particular, we use the Planck 2018 temperature, polarization, and lensing data (using the nuisance-marginalized 'lite' version, as described in \cite{planck_lite_1,planck_lite_4}), and consider either a model of a curved universe ($\Lambda$CDM + $\Omega_k$) or a model with sinusoidal variations of the primordial power spectrum, similar to \cite[Sec 7.1.1]{Planck:2018jri}. For the $\Lambda$CDM baseline model in both cases we adopt the common 6 cosmological parameters $\{\ln(10^{10}A_s), n_s, H_0, \Omega_b h^2, \Omega_\mathrm{cdm} h^2, \tau_\mathrm{reio}\}$ and adopt a single massive neutrino with mass 0.06eV (see \cite{Planck:2018vyg} for a more detailed description of this baseline model).

\begin{figure}
    \centering
    \includegraphics[width=0.8\textwidth]{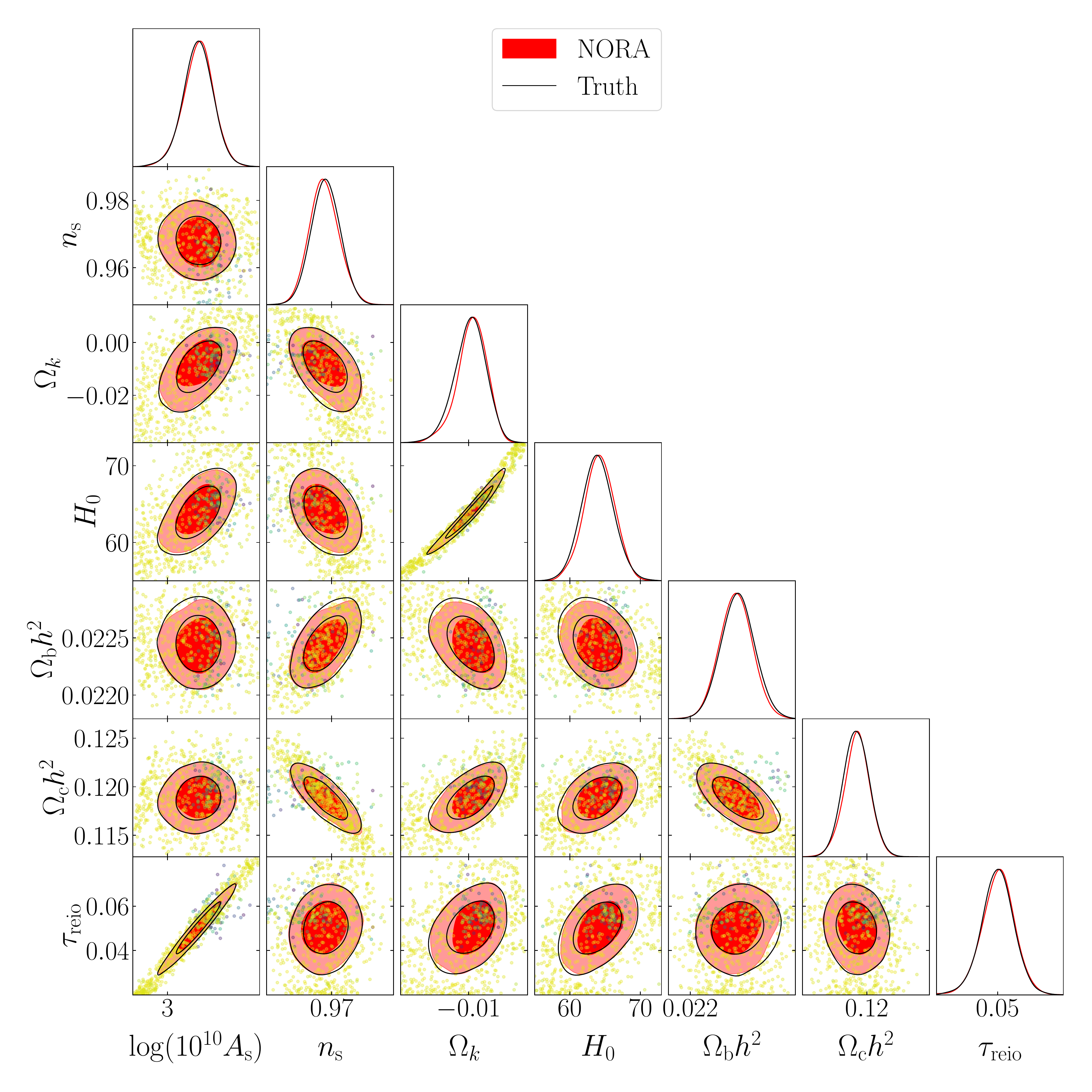}
    \caption{Inference of the cosmological parameters of the Planck 2018 likelihood (Planck lite) with curvature $\Omega_k$ sampled in addition. NORA correctly recovers the contours with only 903 evaluations of the likelihood function.}
    \label{fig:planck_omega_k}
\end{figure}

We show in \Cref{fig:planck_omega_k} the results for a model of a curved universe, an extension of the $\Lambda$CDM model described above with an additional seventh parameter $\Omega_k$ representing the energy density-equivalent of the curvature. It presents a particularly strong degeneracy between the curvature parameter $\Omega_k$ and the Hubble constant $H_0$. We observe that the contours are in good agreement ($D_{\mathrm{KL}}^\mathrm{sym,Gaussian}=0.08$ using the Gaussian approximation of \Cref{app:eq:gaussian_kl}).

\begin{figure}
    \centering
    \includegraphics[width=0.8\textwidth]{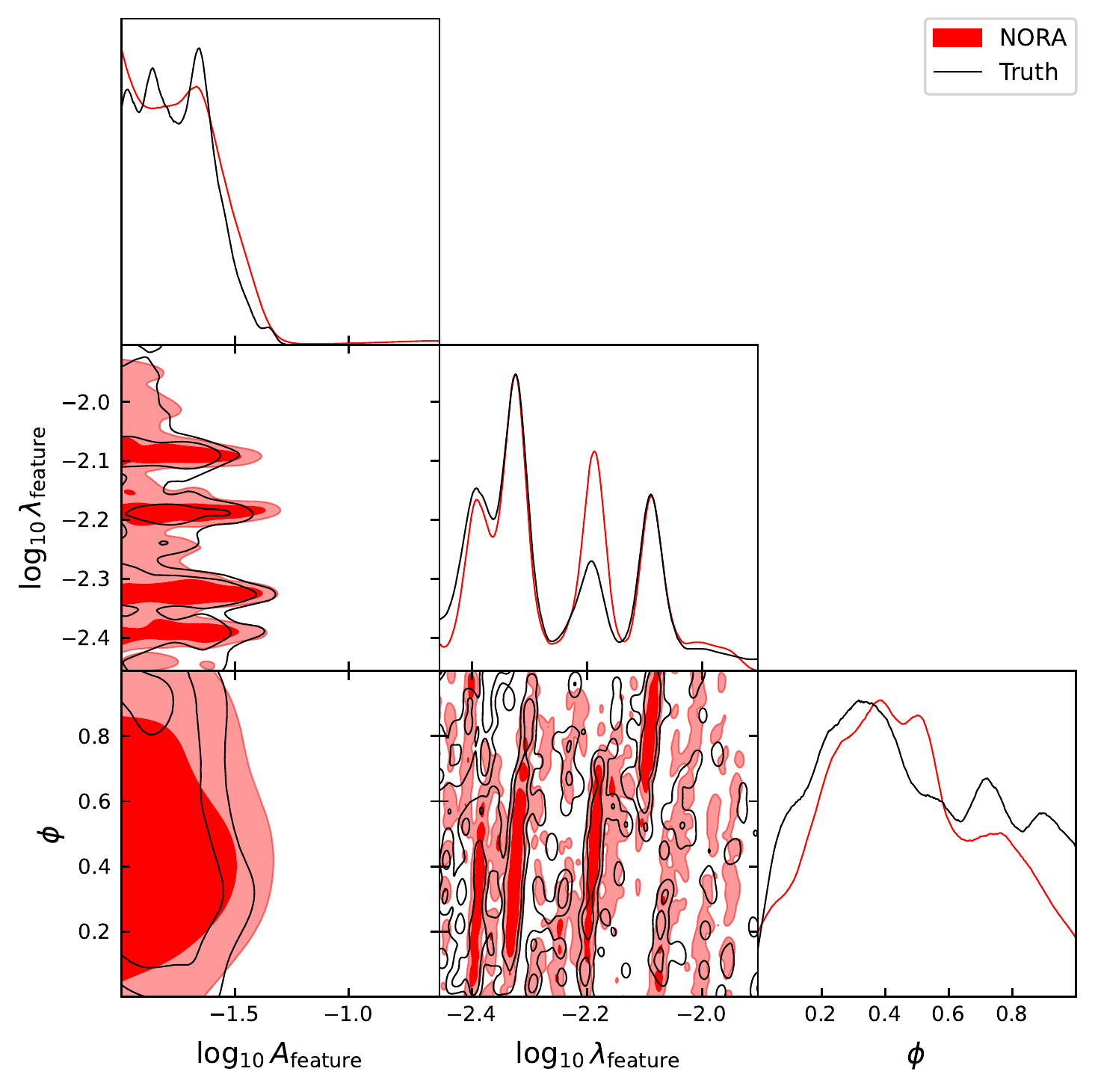}
    \caption{A three dimensional model of an oscillation in the primordial power spectrum of the Planck 2018 CMB sky constrained by NORA, and the reference Nested Sampling results with \texttt{PolyChord}. NORA accurately captures the constraints with a budget of 2000 evaluations, performed in parallel batches of 6 evaluations.}
    \label{fig:planck_sin}
\end{figure}

Next we try to fit a sinusoidal oscillation with three parameters (amplitude, wavelength and phase) to the primordial power spectrum of Planck 2018, fixing the parameters of the $\Lambda$CDM model. This is a low-dimensional problem, but with a highly multi-modal behavior in the frequency and phase of the oscillation, since we are effectively fitting experimental noise. The result can be seen in \Cref{fig:planck_sin}: most of the distribution is well recovered, despite its complexity.

\section{Reproducibility}

The NORA implementation and all scripts required to reproduce the tests will be released after review of this manuscript.
